\documentclass[sigconf, screen]{acmart}

\AtBeginDocument{%
  }

\usepackage[capitalize]{cleveref}  
\setcopyright{acmlicensed}
\copyrightyear{2018}
\acmYear{2018}
\acmDOI{XXXXXXX.XXXXXXX}
\acmConference[Conference acronym 'XX]{Make sure to enter the correct
  conference title from your rights confirmation email}{June 03--05,
  2018}{Woodstock, NY}
\acmISBN{978-1-4503-XXXX-X/2018/06}

\acmSubmissionID{1617}

\begin{document}

\title{FATE: A Prompt-Tuning-Based Semi-Supervised Learning Framework for Extremely Limited Labeled Data}

\author{Hezhao Liu}
\email{23020231154209@stu.xmu.edu.cn}
\authornotemark[1]
\affiliation{%
  \institution{Xiamen University}
  \city{Xiamen}
  \state{Fujian}
  \country{China}
}

\author{Yang Lu}
\email{luyang@xmu.edu.cn}
\affiliation{%
  \institution{Xiamen University}
  \city{Xiamen}
  \state{Fujian}
  \country{China}
}

\author{Mengke Li}
\email{mengkejiajia@hotmail.com}
\affiliation{%
  \institution{Guangdong Laboratory of Artificial Intelligence and Digital Economy (SZ)}
  \city{Shenzhen}
  \state{Guangdong}
  \country{China}
}

\author{Yiqun Zhang}
\email{yqzhang@gdut.edu.cn}
\affiliation{%
 \institution{Guangdong University of Technology}
 \city{Guangzhou}
 \state{Guangdong}
 \country{China}}

\author{Shreyank N Gowda}
\email{shreyank.narayanagowda@nottingham.ac.uk}
\affiliation{%
  \institution{University of Nottingham}
  \city{Nottingham}
  \country{UK}}

\author{Chen Gong}
\email{chen.gong@njust.edu.cn}
\affiliation{%
  \institution{Nanjing University of Science and Technology}
  \city{Nanjing}
  \state{Jiangsu}
  \country{China}}

\author{Hanzi Wang}
\email{Hanzi.Wang@xmu.edu.cn}
\affiliation{%
  \institution{Xiamen University}
  \city{Xiamen}
  \state{Fujian}
  \country{China}
}

\renewcommand{\shortauthors}{Trovato et al.}

\begin{abstract}
Semi-supervised learning (SSL) has achieved significant progress by leveraging both labeled data and unlabeled data. Existing SSL methods overlook a common real-world scenario when labeled data is extremely scarce, potentially as limited as a single labeled sample in the dataset. General SSL approaches struggle to train effectively from scratch under such constraints, while methods utilizing pre-trained models often fail to find an optimal balance between leveraging limited labeled data and abundant unlabeled data. To address this challenge, we propose Firstly Adapt, Then catEgorize (FATE), a novel SSL framework tailored for scenarios with extremely limited labeled data. At its core, the two-stage prompt tuning paradigm FATE exploits unlabeled data to compensate for scarce supervision signals, then transfers to downstream tasks. Concretely, FATE first adapts a pre-trained model to the feature distribution of downstream data using volumes of unlabeled samples in an unsupervised manner. It then applies an SSL method specifically designed for pre-trained models to complete the final classification task. FATE is designed to be compatible with both vision and vision-language pre-trained models. Extensive experiments demonstrate that FATE effectively mitigates challenges arising from the scarcity of labeled samples in SSL, achieving an average performance improvement of 33.74\% across seven benchmarks compared to state-of-the-art SSL methods. Code is available at \href{https://anonymous.4open.science/r/Semi-supervised-learning-BA72}{https://anonymous.4open.science/r/Semi-supervised-learning-BA72}.
\end{abstract}  

\ccsdesc[500]{Computing methodologies~Computer vision problems}
\ccsdesc[500]{Computing methodologies~Computer vision}

\keywords{Semi-Supervised Learning, Prompt Tuning, Pre-trained Model, Few-shot Learning, CLIP, Parameter-Efficient Fine-Tuning}

\maketitle

\begin{figure}[!t]
\begin{center}
\centerline{\includegraphics[width=\columnwidth]{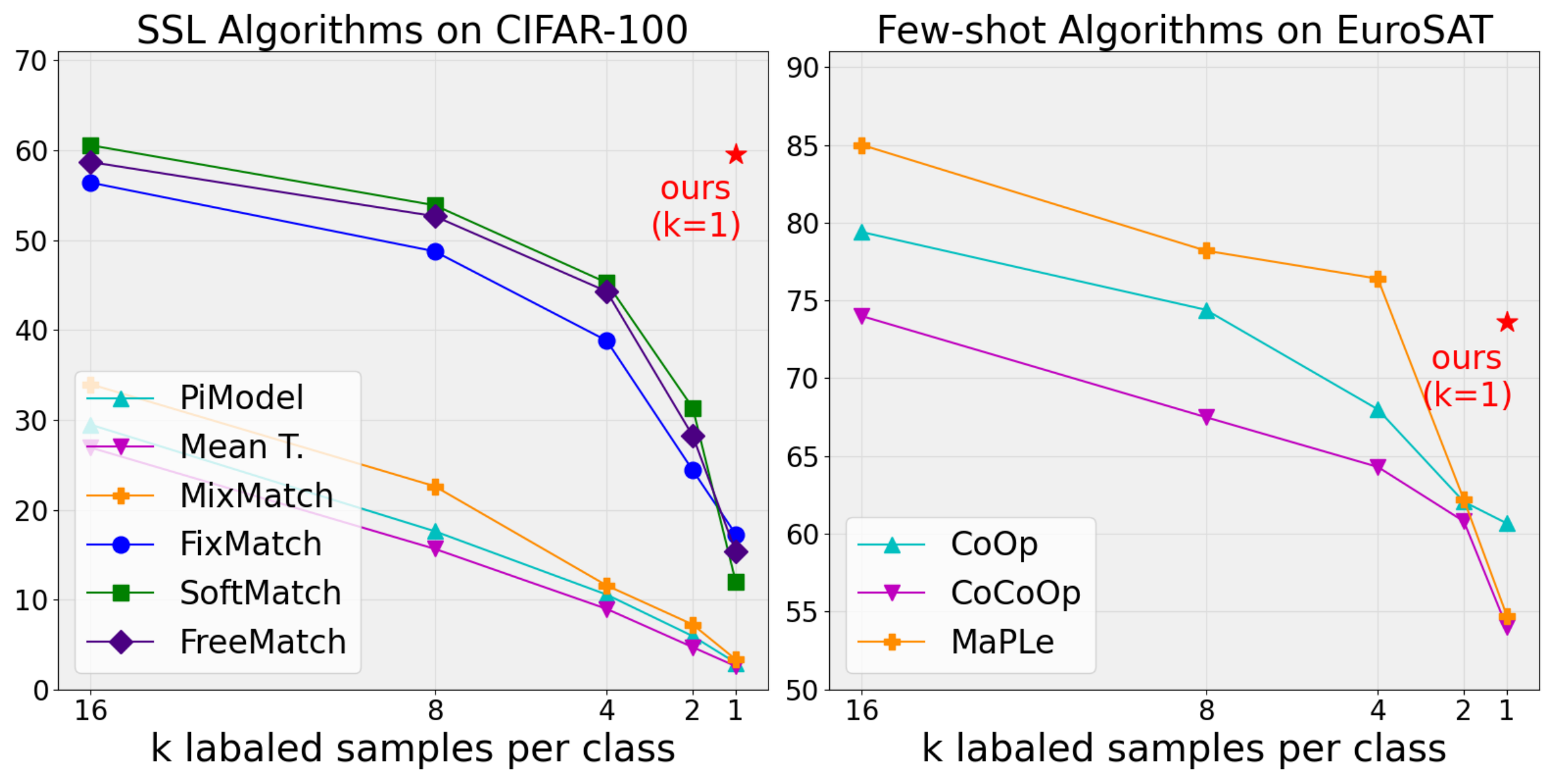}}
\caption{Performance of SSL and few-shot algorithms under different amounts of labeled data $k$ for each class. Generally, the performance degrades as the number of labeled data decreases.}
\label{motivation}
\end{center}
\end{figure}

\section{Introduction}
Semi-supervised learning (SSL) is a machine learning paradigm that aims to leverage both labeled and unlabeled data to improve model performance. Compared to supervised learning, which relies solely on labeled data, SSL can enhance learning efficiency by utilizing additional unlabeled instances \cite{yang2022survey, van2020survey}. Due to its ability to reduce reliance on large-scale labeled datasets, SSL has been successfully applied in various real-world scenarios like medical imaging, natural language processing, and autonomous driving \cite{wang2020focalmix, dai2015semi, hu2022pseudoprop}, making it a crucial research area in the machine learning community.

Existing SSL methods \cite{chen2023softmatch, wang2022freematch, tarvainen2017mean, sajjadi2016regularization} typically assume the availability of a sufficient amount of labeled data, even if the full annotation is not required. However, obtaining even a small fraction of labeled data can be challenging in real-world scenarios. Consider tasks such as medical image annotation or satellite image labeling, which demand substantial human and financial resources. In such cases, datasets may contain only one or two labeled samples while the vast majority remain unlabeled, posing a significant challenge for SSL methods that rely on training a strong feature extractor from scratch. As shown in \cref{motivation}, we evaluate six classical and state-of-the-art (SOTA) SSL algorithms with varying numbers of labeled samples. The results indicate that as labeled data decreases, the performance of these SSL algorithms degrades sharply. For instance, SoftMatch \cite{chen2023softmatch} exhibits nearly a 50\% performance drop when reducing labeled samples per class from 25 to just one. The core challenge lies in training a powerful feature extractor from scratch with extremely limited labeled data. On the other hand, with the advent of pre-trained models, this challenge can be mitigated by leveraging a robust backbone, thereby eliminating the need for training from scratch. A few labeled samples can be used to efficiently fine-tune the pre-trained model, and several methods \cite{jia2022visual, zhou2022learning, chen2022adaptformer, hu2022lora} have achieved promising results. These methods are collectively referred to as Parameter-Efficient Fine-Tuning (PEFT) \cite{lialin2023scaling}. As further demonstrated in \cref{motivation}, our evaluation of three SOTA few-shot PEFT algorithms \cite{zhou2022conditional, zhou2022learning, khattak2023maple} shows that their performance degrades significantly as the number of labeled samples decreases, highlighting their susceptibility to labeled data scarcity. Moreover, these methods lack effective mechanisms to leverage unlabeled data, resulting in substantial information waste.

To fully leverage abundant unlabeled data and mitigate the challenge of extremely scarce labeled data in SSL, we propose Firstly Adapt, Then catEgorize (FATE), a novel prompt-tuning-based SSL framework. The key idea of FATE is to first adapt a pre-trained model to the downstream feature distribution using unsupervised learning, then refine it with limited labeled samples to enhance classification performance. Specifically, FATE leverages unlabeled data for unsupervised adaptation, aligning the model with downstream distributions before incorporating labeled data for final classification. This ensures that the model benefits from the vast amount of unlabeled data while overcoming the limitations of scarce supervision. FATE introduces a flexible two-stage approach that applies to both vision and vision-language pre-trained models, incorporating tailored prompt tuning strategies. We validate the effectiveness of applying FATE on the foundational models of the vision model \cite{dosovitskiy2020image} and the vision language model \cite{radford2021learning}. Extensive experiments demonstrate that FATE outperforms existing SOTA SSL and PEFT methods. Our contributions can be summarized as follows.
\begin{itemize}
  \item We propose FATE, a novel two-stage prompt tuning framework for pre-trained models, specifically designed to address SSL challenges under extremely limited labeled data.
  \item FATE is adaptable to both vision and vision-language pre-trained models, mitigating the transferability limitations of prompt learning between models of different modalities.
  \item The proposed FATE achieved an
  average performance improvement of 33.74\% across seven benchmarks compared to state-of-the-art SSL methods in visual classification tasks.
\end{itemize}

\section{Related Work and Preliminaries}
\subsection{Semi-Supervised Learning}
SSL uses a large amount of unlabeled data and a small number of labeled data to build models. SSL methods can be roughly divided into two categories \cite{yang2022survey}: consistency regularization \cite{sajjadi2016regularization, tarvainen2017mean, xie2020unsupervised} and pseudo-labeling \cite{dong2018tri, xie2020self, chen2020big}. Consistency regularization methods are often based on multi-model architectures, allowing models to learn from each other and adding regularization terms to the input, model parameters, and training strategies to obtain more robust feature extractors \cite{yang2022survey}. Pseudo-labeling methods extend labeled datasets by assigning high-confidence pseudo-labels to unlabeled data to solve the problem of insufficient labeled data. Currently, many methods integrating the above two ideas have been proposed \cite{chen2023softmatch, wang2022freematch, chen2023boosting, tan2023otmatch}, with FixMatch \cite{sohn2020fixmatch} being one of the most representative. The key innovation of FixMatch comes from combining these two ingredients and using separate weak and strong augmentations in the consistency regularization approach. Given an unlabeled instance, only when the model predicts a high-confidence label can the predicted pseudo-label be identified as ground truth. Formally, to define a $Y$-class SSL classification problem, we let $\mathcal{X} = \{({x_b}, {y_b}) \mid b \in (1,\dots, B)\}$ be a batch of $B$ labeled sample, where $x_b$ is the training sample with its one-hot label $y_b$. Let $\mathcal{U} = \{{u_b}\mid b \in (1,\dots, \mu B)\}$ be a batch of $\mu B$ unlabeled samples, where $\mu$ is a fixed parameter that determines the ratio of $\mathcal{X}$ to $\mathcal{U}$. 

\subsection{Prompt Learning for Pre-trained Model}
In recent years, prompt tuning, a fine-tuning method for pre-trained models \cite{radford2018improving, devlin2018bert} in the field of NLP, has received widespread attention. Prompt tuning requires adapting the pre-trained model to downstream tasks without updating its parameters. Downstream data refers to the entire dataset of our specific downstream task. Jia \cite{jia2022visual} first proposed \textit{Visual Prompt Tuning} (VPT) based on the architecture of Vision Transformer (ViT) \cite{dosovitskiy2020image}, which adds learnable embedding vectors to the image patch embedding vectors in various ways. During training, the parameters of the ViT remain unchanged, and only the parameters of the prompt are updated. It is only necessary to fix the prompt when it comes to the inference phase \cite{jia2022visual, zhu2023visual, bar2022visual}. Specifically, VPT \cite{jia2022visual} adds learnable embedding vectors to the image patch embedding vector of ViT \cite{dosovitskiy2020image}. For a ViT model denoted as $V$, an input image $I$ is cut into $m$ fixed-sized patches. Each patch is embedded into $d$-dimensional latent space with positional encoding. The process of cutting and embedding can be denoted as $\mathcal{E}$. The collection of input image patch embeddings $E$ is: $E = \mathcal{E}(I)$, where $E \in \mathbb{R}^{m \times d}$.
Together with an extra frozen classification token and a group of learnable tokens $P\in \mathbb{R}^{n\times d}$, the whole ViT is formulated as:
\begin{align}
    [x_{cls}^{out}; E^{out}; P^{out}] = V([x_{cls}^{in}; E^{in}; P^{in}]),
\end{align}
where $x_{cls}$ denotes the classification token, $n$ denotes length of the group of prompts. Only $P$ is updated while $V$ and $x_{cls}$ are kept frozen. For brevity, the symbols $in$ and $out$ below will be omitted. Zhou \cite{zhou2022learning} proposed \textit{Context Optimization}, known as CoOp, a simple approach specifically for adapting CLIP-like vision-language models for downstream image recognition. CoOp models a prompt's context words with learnable vectors while the pre-trained parameters are kept fixed. Concretely, The original CLIP \cite{radford2021learning} works in the following way: for a textual encoder denoted as $T$ of Transformer \cite{vaswani2017attention}, a fixed sequence of words, such as ``a photo of a $\mathtt{[CLASS]}$." will be converted with byte pair encoding representation \cite{sennrich2015neural} and encompassed at a fixed length $l$ to get input embedding vectors $S\in\mathbb{R}^{l \times d}$. the process can be denoted as $f = T(S)$, where $f\in\mathbb{R}^d$ represents the text feature of $S$. CLIP calculates the similarity between the text features $f$ and image features from a visual encoder. The class with the highest similarity is the predicted class of the image sample. CoOp \cite{zhou2022learning} replaces the embedding vectors of fixed prompt ``a photo of a" with a set of learnable tokens $P\in \mathbb{R}^{n\times d}$ to get new input embedding vectors $S'$ and new text feature $f'$. Only $P$ are updated during training and $P$ are fixed during inference.

\section{Method}

\subsection{Motivation}
As classical SSL algorithms struggle to train an effective backbone from scratch under conditions of extreme scarcity of labeled data, we leverage pre-trained models to obtain a powerful feature extractor and fine-tune it, thereby avoiding training from scratch. However, directly fine-tuning the pre-trained model with only labeled data will lead to serious information waste because there is no mechanism to utilize unlabeled data. We aim to develop a method with a mechanism that effectively leverages unlabeled data while retaining some pre-trained knowledge from the pre-trained model. Compared to other PEFT approaches \cite{hu2022lora, houlsby2019parameter}, prompt tuning does not require modifications to the backbone network and achieves high computational efficiency by optimizing only the input-related parameters to learn task-specific knowledge. Therefore, the prompt tuning method aligns perfectly with our goal of utilizing unlabeled data to adapt the pre-trained model while preserving some pre-training knowledge. 

In \cref{framework}, we provide a detailed introduction to the proposed FATE framework, followed by its implementation on the vision pre-trained model in \cref{vit-imple} and the vision-language pre-trained model in \cref{clip-imple}, respectively.
\begin{figure*}[!t]
\centering
{\includegraphics[width=\textwidth]{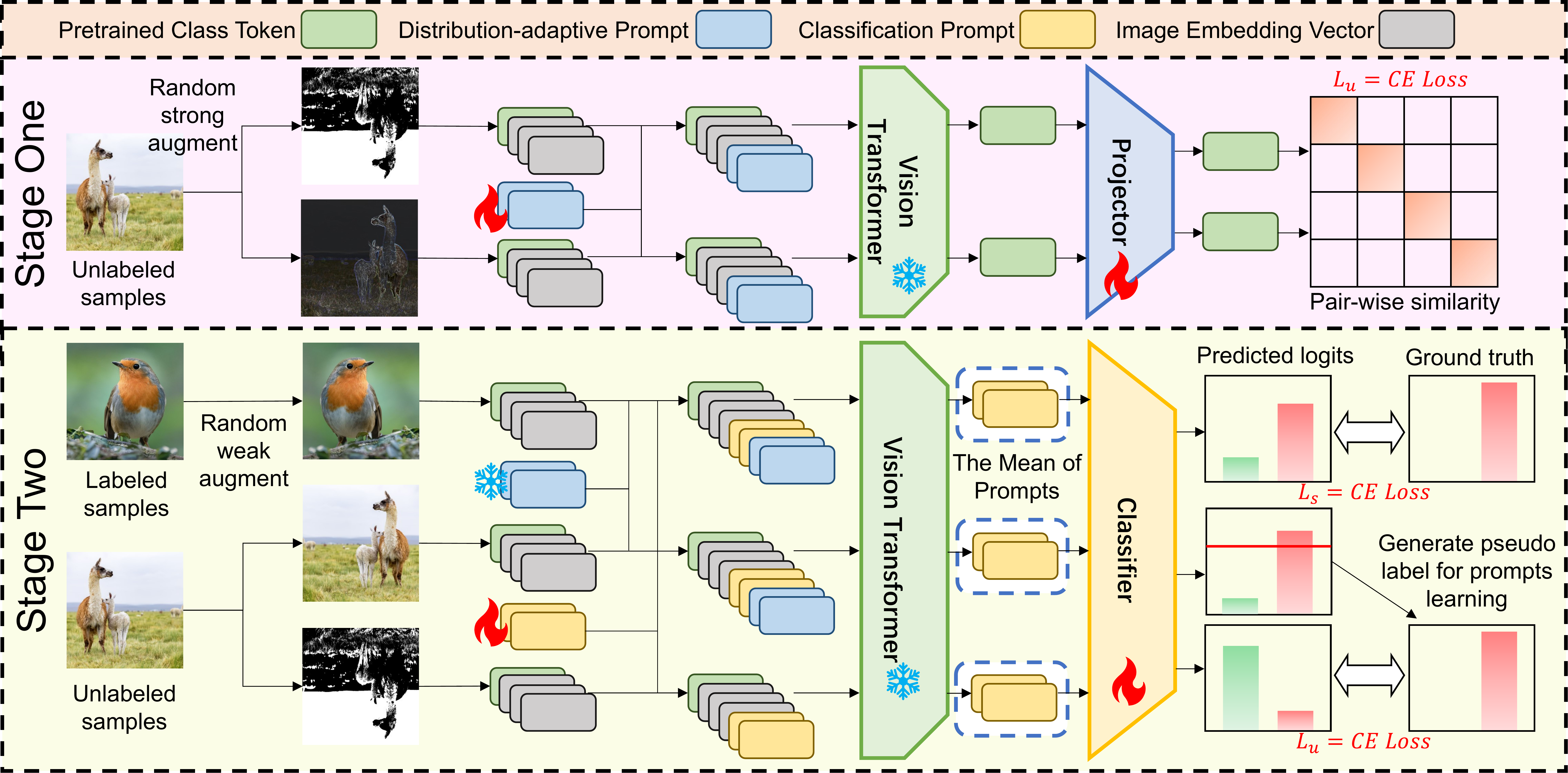}}
\caption{Implementation of FATE on the vision model. Firstly, we concatenate the DP to the embedding vectors of the unlabeled samples and optimize it with contrastive learning loss. Then we modify FixMatch by first concatenating learnable CP to all embedding vectors, fixing the DP just learned, and concatenating it to the branches of the weak augmented views of both labeled and unlabeled samples. Finally, the mean of the CP is used as the classification feature.}
\label{vit}
\end{figure*}


\subsection{FATE Framework}
\label{framework}
To implement the aforementioned idea, we propose a two-stage prompt tuning framework for pre-trained models named FATE. In the first stage, we leverage a large amount of unlabeled data to adapt the model to the downstream data's feature distribution, referred to as the \textit{Adaptation Stage}. In the second stage, we utilize the entire downstream dataset to perform the final SSL classification task, referred to as the \textit{Classification Stage}.
\paragraph{Adaptation Stage.} Before tuning a pre-trained model with a few labeled data, FATE explores the feature distribution of a large amount of unlabeled data and stores this distribution information into the prompt tokens through prompt learning. Formally, for a frozen pre-trained model $M$, we first use unlabeled samples $\mathcal{U}$ to update a group of prompt tokens through unsupervised learning methods with $\mathcal{L}_u$. This set of prompts is designed to learn the overall feature distribution of downstream data, so we refer to the prompts as Distribution-adaptive Prompts (DP), denoted as $P_d$. The process can be defined as:
\begin{align}
    P_d^* = \underset{P_d}{\arg\min}\frac{1}{\mu B}\sum^{\mu B}_{b=1}\mathcal{L}_{u}(M(u_b;P_d)), \label{adaptstageloss}
\end{align}
where $P_d^*$ is the optimized DP. The model’s understanding of downstream data is improved by incorporating $P_d^*$ during prediction. $P_d^*$ remains fixed and is used during both the Classification Stage and final inference.


\paragraph{Classification Stage.} Since unlabeled data was only used to help the model adapt to the feature of downstream data, generally rather than directly for the classification task, the vast amount of feature information provided by unlabeled data is not directly exploited for classification. Therefore, we again leverage unlabeled data along with the supervised signal from a small amount of labeled data to complete the classification task instead of simply fine-tuning the model with labeled samples. This approach allows unlabeled data to fully contribute to adapting the model at a coarse level and improving its classification performance at a fine-grained level.

Formally, we freeze the DP just learned and optimize a new set of prompts named Classification Prompt (CP), denoted as $P_c$, to perform the final classification task. We further modify FixMatch \cite{sohn2020fixmatch} through prompt learning to adapt it to the pre-trained model architecture. For each unlabeled sample $u_b$ in $\mathcal{U}$, frozen DP is incorporated when the weak augmentation view of $u_b$ is input into the model. During the forward propagation of data features, incorporating DP enhances the model's understanding of downstream data. Thus, the model's predicted class distributions for both the strong augmented view $q_s$ and weak augmented view $q_w$ can be denoted as:
\begin{align}
    q_s &= M(\mathcal{A}(u_b); P_c), \label{strong}\\
    q_w &= M(\mathcal{W}(u_b); [P_d^*; P_c]), \label{weak}
\end{align}
where $\mathcal{A}(\cdot)$ and $\mathcal{W}(\cdot)$ represent the strong and the weak augmentation operation to the sample, respectively. The unsupervised loss of unlabeled data can be defined as: 
\begin{align}
    \mathcal{L}_u = \frac{1}{\mu B}\sum^{\mu B}_{b=1} 1_{[\max(q_w) \geq \theta]} \mathcal{H}(\hat{y}_b, q_s) \label{unsupervised},
\end{align}
where $\hat{y}_b=\arg\max(q_w)$ represents the one-hot pseudo-label of the unlabeled sample $u_b$, and $\hat{y}_b$ is generated by the predicted class distribution of $\mathcal{W}(u_b)$. $\theta$ is a hyperparameter representing the threshold above to retain a pseudo-label, and $\mathcal{H}(\cdot,\cdot)$ represents the cross-entropy loss function.
For each labeled sample $x_b$ in $\mathcal{X}$, DP is incorporated again when $\mathcal{W}(x_b)$ is input into the model. The corresponding predicted class distribution and the supervised loss of labeled data are:
\begin{align}
    q_l &= M(\mathcal{W}(x_b); P_d^*; P_c), \label{label} \\
    \mathcal{L}_s &= \frac{1}{B}\sum^B_{b=1}\mathcal{H}(y_b, q_l) \label{supervised}.
\end{align}
This process of FATE can be defined as:
\begin{align}
    P_c^* = \underset{P_c}{\arg\min}(\mathcal{L}_s + \lambda\mathcal{L}_u)\label{classstageloss}, 
\end{align}
where $\lambda$ is a hyperparameter denoting the relative weight of the unsupervised loss and $P_c^*$ represents the optimized CP. During inference, both sets of prompts DP and CP are fixed and concatenated with the input embedding vectors to obtain the prediction value.

\begin{figure*}
\centering
{\includegraphics[width=\textwidth]{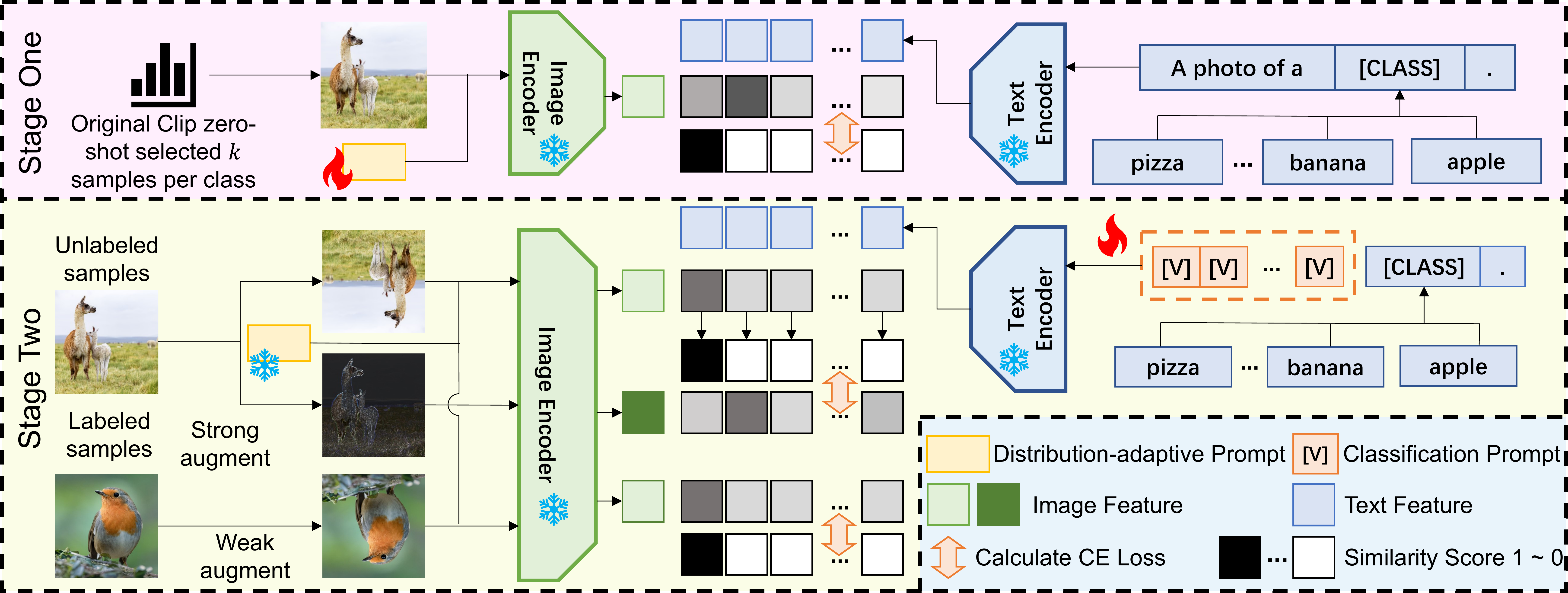}}
\caption{Implementation of FATE on the vision-language model. We use the original CLIP's zero-shot capability to pseudo-label the unlabeled data, selecting the top-$k$ samples with the highest predicted values for each class and training the DP for the visual encoder with the pseudo-labeled samples. Then we fix the DP just learned and design CP at the textual encoder side, optimizing it using \cref{classstageloss}.}
\label{clip}
\end{figure*}

\subsection{Implementation}
\noindent\textbf{Vision Model}
\label{vit-imple}
For vision models, the focus of FATE implementation should be on the feature level of the data.
\paragraph{Adaptation Stage.} For the pre-trained vision model like ViT that requires a classifier for classification, we can use contrastive learning to enable DP to recognize different augmented views of the same sample. This can achieve the purpose of utilizing the unlabeled data to let the model learn the overall feature information of the downstream data mentioned in \cref{framework}. For each unlabeled sample $u_b$ in $\mathcal{U}$, as shown in \cref{vit}, we make two random strong augmented views $\mathcal{A}(u_b)_1$ and $\mathcal{A}(u_b)_2$ where $\mathcal{A}(u_b)_1 \ne \mathcal{A}(u_b)_2$, then cut and map them into embedding vectors: $E_{s}^1 = \mathcal{E}(\mathcal{A}(u_b)_1)$ and $E_{s}^2 = \mathcal{E}(\mathcal{A}(u_b)_2)$. We concatenate DP on the two sets of input embedding vectors $E_s^1$ and $E_s^2$ and then add a frozen pre-trained class token $x_{cls}$. Two final input vector groups from $\mathcal{A}(u_b)_1$ and $\mathcal{A}(u_b)_2$ and their corresponding output vectors are:
\begin{align}
    [x_{cls}^{1}; P_d^{1};E_{s}^{1}] = V([x_{cls}; P_d;E_{s}^{1}]),\\
    [x_{cls}^{2}; P_d^{2};E_{s}^{2}] = V([x_{cls}; P_d;E_{s}^{2}]),
\end{align}
Note that the class token and DP in the two input vector sets are the same, but their values differ after ViT mapping due to the distinct embedding vector groups of the strong augmented views ($E^1_s \ne E^2_s$). We take the two mapped $x_{cls}^1$ and $x_{cls}^2$ and project them into a low-dimensional space using a projector $J$ to get two projected class tokens $x_{cls}^1$ and $x_{cls}^2$. For batch unlabeled samples, each unlabeled sample $u_i$ in $\mathcal{U}$ is corresponding two projected class tokens $x_{cls}^{i, 1}, x_{cls}^{i, 2}$ where $i\in\mu B$. The distance between the two class tokens $x_{cls}^{i, 1}$ and $x_{cls}^{i, 2}$ obtained by augmenting the two views with the same sample $u_i$ should be shortened, and the distance between the class tokens $x_{cls}^{i, *}, x_{cls}^{j, *}$ of each view augmented with different samples should be further, where $i,j\in\mu B, i \ne j$ and $*$ represents any of the two augmented views. The contrastive loss of the batch of unlabeled data, also serving as the unsupervised loss \cref{adaptstageloss} used to optimize DP mentioned in \cref{framework} can be denoted as: 
\begin{align}
    \mathcal{L}_{u} = -\sum_{i=1}^{\mu B}\log\frac{\exp(\langle x_{cls}^{i,1}, x_{cls}^{i,2}\rangle / \tau)}{\sum_{k=1}^{2\mu B}1_{[k\ne i]}\exp{(\langle x_{cls}^{k,*},x_{cls}^{i,*}\rangle/\tau)}}, \label{contrastive}
\end{align}
where $\tau$ is a scalar temperature hyperparameter adjusting the distribution of similarity scores and $\langle \cdot, \cdot \rangle$ represents the cosine similarity. Only $P_d$ and $J$ are updated in this process. 

\paragraph{Classification Stage.} After the model has completed the adaptation to the downstream data by optimizing DP, we leverage all data to complete the classification task. In contrast to the Adaptation Stage, the pre-trained class token is treated as a feature vector rather than the classification feature, as $x_{cls}$ has been used for feature contrastive learning instead of classifying. For each unlabeled sample or labeled sample, we can obtain its embedding vectors and corresponding input vector groups based on the method proposed in \cref{framework}. After the input vector groups are mapped by ViT, the mean of $P_c$ will serve as the high-dimensional feature for classification, which is then fed into a classifier $C$ to obtain $q_s$, $q_w$ and $q_l$ based on \cref{strong}, \cref{weak} and \cref{label}. The supervised loss $\mathcal{L}_s$ and the unsupervised loss $\mathcal{L}_u$ can be calculated using \cref{supervised} and \cref{unsupervised}. Then update $P_c$ and $C$ using \cref{classstageloss}.  

\begin{table*}[t]
\caption{Accuracies on seven datasets for SOTA methods in SSL. The backbone used in the upper part is Wide-Resnet \cite{zagoruyko2016wide}, and the backbone used in the lower part is ViT. Bold indicates the best performance.}
\label{vit-result}
\begin{tabular}{lcccccccc}
\toprule
   Method  & CIFAR-10 & CIFAR-100 & ImageNet100 & Semi-iNat & SVHN & SUN397 & CUB200 & Avg.\\
\midrule
Pseudo-label \cite{arazo2020pseudo}   & 19.07& 3.43& 2.98 & 1.87& 18.69& 1.43& 2.37 &7.12 \\
FixMatch \cite{sohn2020fixmatch} & 17.36& 14.59& 5.06& 1.76& 11.05& 2.48& 3.53 &7.98\\
MixMatch \cite{berthelot2019mixmatch}  & 25.37& 3.90& 2.10 & 1.95& 20.36& 1.85& 1.94&8.21 \\
SoftMatch \cite{chen2023softmatch}  & 49.62& 9.30& 6.44&3.26& 16.56& 6.05& 4.52  &13.68  \\
FullMatch \cite{chen2023boosting}   & 26.25& 20.77& 5.96 & 1.88& 12.35& 2.76& 5.73 &10.81\\
FreeMatch \cite{wang2022freematch}    & 75.65& 15.69&7.04& 2.81& 18.25& 2.94& 5.17 &18.22\\
\midrule
Pseudo-label \cite{arazo2020pseudo}   & 19.00& 4.97&19.42& 2.24& 12.07& 3.24& 2.45 &9.06\\
FixMatch \cite{sohn2020fixmatch}& 26.59& 10.02 & 12.56 & 2.44& 9.72& 4.58& 13.78 &11.38\\
SoftMatch \cite{chen2023softmatch}  & 29.15& 5.85& 6.22 & 2.47& 10.11& 1.86& 3.23 & 8.41\\
Fine-Tuning Classifier (sup)    & 52.48& 34.17& 74.72 & 20.62& 16.22& 31.93& 41.11  & 38.75  \\
VPT (sup) \cite{jia2022visual}   & 51.54& 34.69& 74.20 & 20.70& 15.90& 30.97& 40.59 & 38.37\\
FATE (Ours)      & \textbf{85.89}& \textbf{59.55}& \textbf{76.48} & \textbf{23.86}& \textbf{20.97}& \textbf{40.03}& \textbf{56.96} & \textbf{51.96}\\
\bottomrule
\end{tabular}
\end{table*}

\noindent\textbf{Vision-Language Model Implementation}
\label{clip-imple}
For vision-language models, we should prioritize the extraction of semantic features from the data.

\paragraph{Adaptation Stage.} For the vision-language model, such as CLIP, we can utilize its strong zero-shot capability to generate pseudo-labels for the unlabeled samples. Following \cite{menghini2023enhancing}, we then select samples with high predicted logits to represent the feature distribution of the downstream data because the features of these samples are quite distinct and easy to classify. Based on the idea mentioned in \cref{framework}, we can let DP learn the feature distribution of the downstream data with these pseudo-labeled samples. Specifically, as shown in \cref{clip}, we first pseudo-label all unlabeled data $\mathcal{U}$ using the original CLIP, and then divide the unlabeled samples according to the pseudo-labeled classes. We take the $k$ samples with the highest predicted logits (similarity score) in each class to get a set of samples with high confidence pseudo labels: $\hat{\mathcal{X}} = \{(\hat{u}_c, \hat{y}_c)\mid c \in(1,\dots, Y\times k)\}$. $\hat{\mathcal{X}}$ can be recognized as a representative of the entire downstream unlabeled data's feature distribution. DP is still done on the visual encoder $V$. We use each sample $\hat{u}_c$ in $\hat{\mathcal{X}}$ to patch and embed to form embedding vectors $E_v = \mathcal{E}(\hat{u}_c)$, and splice the DP into $E_v$ and input them together into the frozen visual encoder $V$. At the same time, the class name prompts of the samples ($Y$ sentences in the format of ``a photo of a $\mathtt{[CLASS]}$") are also embedded into vectors $S$, and $S$ are input into the frozen textual encoder $T$. The process of extracting image and text features can be defined as:
\begin{align}
    &[x_{cls}; P_d; E_v] = V([x_{cls}; P_d; E_v]), \\
    &f = T(S),  S\in\mathbb{R}^{Y\times l\times d}, \quad f\in\mathbb{R}^{Y\times d}.
\end{align}
Then calculate the cosine similarity between text features $f$ and image features $x_{cls}$ to get the classification distribution, and calculate the unsupervised loss \cref{adaptstageloss} mentioned in \cref{framework} with the one-hot pseudo-label $\hat{y}_c$ of the sample $u_c$ to update $P_d$:
\begin{align}
    \mathcal{L}_{u}=\frac{1}{k \times Y}\sum_{c=1}^{k \times Y}\mathcal{H}(\hat{y}_c, \langle x_{cls},f\rangle).
\end{align}

\paragraph{Classification Stage.} Following CLIP's adaptation, we replicated our implementation on ViT. However, the difference is that our DP is only modified on the visual side and is not aligned with the textual side. To ensure the consistency between the two modalities encoders, our CP is designed on the textual side according to the prompt design method of CoOp \cite{zhou2022learning} and $P_c$ replaces the embedding vectors of ``a photo of a". The final input embedding vectors $S'$ and the text features $f'$ can be defined as:
\begin{align}
    &S' = \big\{[P_c;\texttt{[CLASS]}_i] \mid i \in( 1,\dots,Y)\big\},
    \\&f'=T(S'),\qquad S'\in\mathbb{R}^{Y\times l\times d}, f'\in\mathbb{R}^{Y\times d}.
\end{align}
For each unlabeled sample or labeled sample, we can get its corresponding class tokens mapped by ViT. The operation is based on \cref{framework} and follows the approach we used on ViT without CP. We calculate the similarity between $x_{cls}$ and $f'$ to obtain the logits $q_s$, $q_w$ and $q_l$ based on \cref{strong}, \cref{weak} and \cref{label}. $\mathcal{L}_s$ and $\mathcal{L}_u$ can be calculated using \cref{supervised} and \cref{unsupervised}, respectively. Finally, we only update $P_c$ using \cref{classstageloss}.


\section{Experiment}
\begin{figure*}[!t]
\centering
{\includegraphics[width=\textwidth]{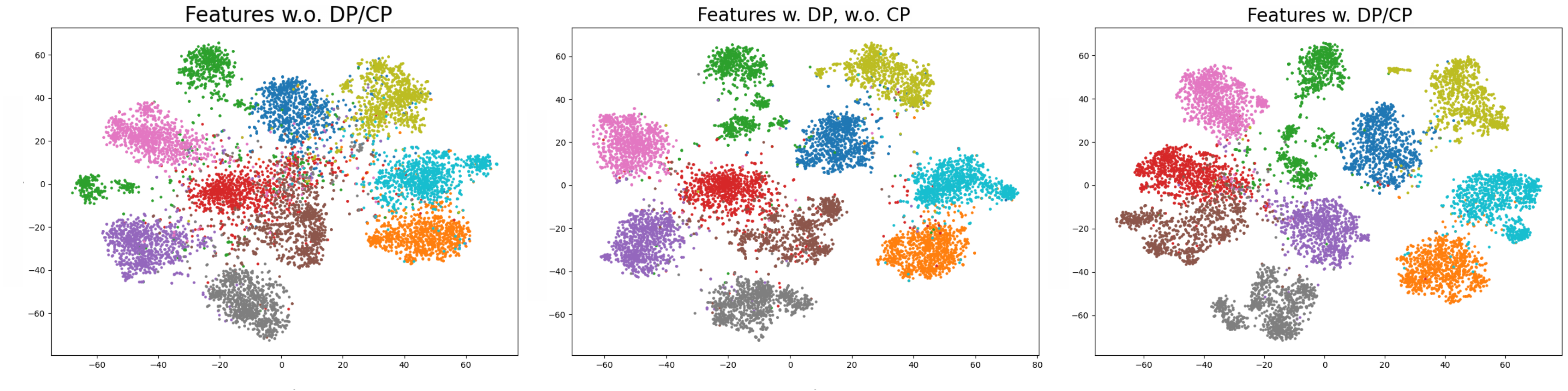}}
\caption{The t-SNE visualization of FATE's implementation on the vision model for the CIFAR-10 test set. With the inclusion of DP/CP, the features of data points belonging to the same class cluster together, whereas the original features without DP/CP remain dispersed.}
\label{t-sne}
\end{figure*}




\subsection{Datasets and Experimental Settings}
\textbf{Basic Settings:} All experiments are conducted through PyTorch \cite{paszke2019pytorch} framework on Ubuntu-20.04.6 with one single GeForce RTX 3090 GPU. To construct training datasets for experiments, we randomly sample \textbf{one labeled instance} from each class in the original dataset, and the collection of these samples is used as a labeled training dataset. The remaining samples in the original dataset are used as unlabeled examples. In all implementations, strong augmentation strategies for unlabeled samples are Random Augmentation \cite{cubuk2020randaugment} including Cutout \cite{devries2017improved}. The relative weight of the unsupervised loss $\lambda$ is set to 1. During training, we randomly sample batches of data from the labeled and unlabeled sets. The average results over three runs with different random seeds are reported.

\textbf{Vision Model Setting:} We use the vit-base-patch16-224 pre-trained model published on hugging face from TIMM deep learning library \cite{steiner2021train, dosovitskiy2020image, rw2019timm} as the feature backbone and evaluate the implementation of FATE on the vision model on seven visual classification benchmarks: (1) CIFAR-10 \cite{krizhevsky2009learning}, (2) CIFAR-100, (3) ImageNet100 \cite{deng2009imagenet}, (4) Semi-iNat \cite{su2021semi}, (5) SVHN \cite{netzer2011reading}, (6) SUN397 \cite{xiao2010sun}, (7) CUB200 \cite{WahCUB_200_2011}. We set $\mu=1$, $B=32$, $\tau=0.5$ and $\theta=0.95$ for training. The training process is done with SGD \cite{ruder2016overview} and an initial learning rate of 0.03, decayed by the cosine annealing rule. Both the length of DP and CP are fixed to 12. The training epoch is set to 10 in the Adaptation Stage and 50 in the Classification Stage. Experiments are implemented based on the SSL PyTorch framework USB \cite{wang2022usb}.

\textbf{Vision-Language Model Setting:} We use the ViT-B/16 CLIP pre-trained model published by OpenAI as the feature backbone and evaluate the implementation of FATE on the vision-language model on ten visual classification benchmarks: (1) StanfordCars \cite{Krause_2013_ICCV_Workshops}, (2) SUN397 \cite{xiao2010sun}, (3) Food101 \cite{bossard14}, (4) OxfordPets \cite{parkhi12a}, (5) DTD \cite{cimpoi14describing}, (6) OxfordFlowers \cite{Nilsback08}, (7) FGVCAircraft \cite{maji2013fine}, (8) Caltech101 \cite{fei2007learning}, (9) EuroSAT \cite{helber2019eurosat}, (10) UCF101 \cite{peng2018two}. We set $\mu=16$, $B=4$, $\tau=0.5$ and $\theta=0.95$ for training. Training is done with SGD \cite{ruder2016overview} and an initial learning rate of 0.1 when adapting and 0.0025 when classifying. The length of DP is fixed to 12 while CP is 16. The training epoch is set to 20 in the Adaptation Stage, and 20 in the Classification Stage. The results when $k=16$ are shown, and the influence of $k$ will be discussed in \cref{k-value}. Experiments are implemented based on the PyTorch framework Dassl \cite{zhou2021domain, zhou2022domain}.

\begin{figure*}[!t]
\centering
{\includegraphics[width=\textwidth]{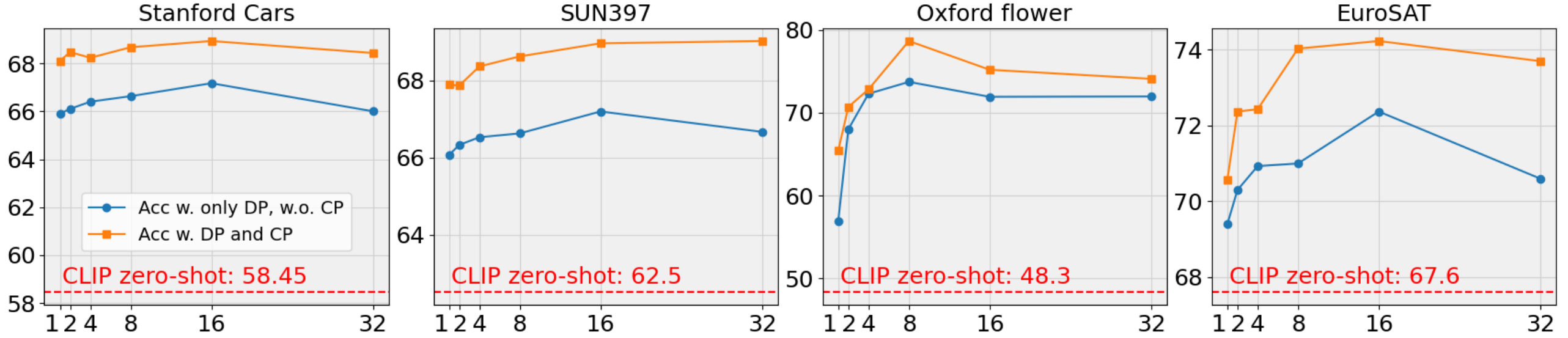}}
\caption{Test accuracy of FATE implemented on the vision language model on four datasets with different $k$. We performed ablation experiments on DP and CP. The model performance with DP is already higher than the original CLIP zero-shot, and with CP, the model performance can be further improved. As $k$ increases to a certain extent, the model's performance tends to converge or decrease. }
\label{clip-k}
\end{figure*}

\subsection{Main Results}
We compared the performance of FATE on the vision model with the current SOTA SSL methods, and the results are shown in \cref{vit-result}. It can be seen that FATE achieves SOTA performance on all seven benchmarks, and FATE can improve the test accuracy by 4.41\% to 50.25\% compared with SoftMatch. FATE's accuracy improved by 3.16\% to 34.35\% compared to directly fine-tuning the vision model with labeled data. This is because the semantic information from only one labeled sample is insufficient to train a strong backbone from scratch. Directly replacing these methods with the vision model and fine-tuning is not feasible. After all, the vision model needs more label data to fine-tune from scratch \cite{dosovitskiy2020image}. Directly fixing the vision model and only fine-tuning a classifier with one labeled sample also has poor performance due to overfitting.

\subsection{Ablation Study on Vision Model}
To prove the role of DP and CP, we conducted an ablation experiment on the vision model. We take Fixmatch with VPT (use Fixmatch to optimize VPT parameters) as the basic method and gradually add modules. The ablation results are shown in \cref{vit-ablation}, and we can see that DP plays an important role in improving the performance of FATE while the inclusion of CP further improved the experimental results. Furthermore, to visually observe the distribution of classification features, we construct 2D t-SNE visualizations of the classification features of the CIFAR-10 test set, adding DP/CP or not. As shown in \cref{t-sne}, the original data feature points would be highly scattered without DP and CP. The above experimental results demonstrate that the DP, optimized through contrastive learning, captures the feature distribution of downstream data. When the model incorporates the DP for classification tasks, it enhances its understanding of downstream data. Meanwhile, CP refines class boundaries, resulting in compact and distinct clusters, further enhancing the model's performance. 

\begin{table}
\caption{The results for the ablation study on the vision model. If CP is not selected, the class token is used as the classification feature.}
\label{vit-ablation}
\begin{tabular}{cccccc}
\toprule
  DP & CP  & CIFAR-100 & SVHN & SUN397 & CUB200 \\
\midrule
 $\times$ &$\times$ & 47.61& 17.09& 35.22& 47.78 \\
 $\surd$& $\times$& 58.35& 17.71& \textbf{40.29}& 54.41\\
 $\times$& $\surd$ & 48.14& 20.21& 37.39& 50.24 \\
 \midrule
 $\surd$&$\surd$&  \textbf{59.55}& \textbf{20.97}& 40.03& \textbf{56.96}  \\
\bottomrule
\end{tabular}
\end{table}

\begin{table}
\caption{The results for fine-tuning a classifier (F C) on the vision model with/without DP using one labeled data.}
\label{vit-more-1}
\begin{tabular}{lccccccc}
\toprule
  &  CIFAR-100 &  SVHN & SUN397 & CUB200 \\
\midrule
 F C w.o. DP&  34.17& 16.22 &31.93& 41.11 \\
F C w. noisy DP& 33.68& 14.37 &26.76& 40.08\\
 \midrule
F C w. DP&  \textbf{40.61}& \textbf{17.09} &\textbf{35.25}& \textbf{46.92}  \\
\bottomrule
\end{tabular}
\end{table}

\begin{table*}[t]
\caption{Accuracies on ten datasets for SOTA methods in other PEFT strategies and FATE's implementation on the Vision-Language model. Underlined values indicate suboptimal performance. Aside from GLoRA, CoOp, CoCoOp, MaPLe, and our experiments, other results are taken from \cite{zanella2024low}.}
\label{PEFT}
\begin{tabular}{lccccccccccc}
\toprule
    Method & Cars & SUN & Food & Pets & DTD & Flowers & Aircraft & Caltech & EuroSAT & UCF & Avg. \\
\midrule
CLIP \cite{radford2021learning}   & 58.5& 62.5& {85.9}& 89.0& 44.1 &70.7& 24.8& 93.3& 48.3& 67.6 &64.47 \\
\midrule
CoOp \cite{zhou2022learning} & 67.5& 66.8& 84.2& 90.4& 49.7& {78.6}& 16.5& 93.3& {60.7}& {72.0}&67.96\\
CoCoOp \cite{zhou2022conditional}  & {68.3}& \underline{68.6}& 85.5& {91.9}& {51.5}& 74.7& 27.6& {94.2}& 54.0& 71.2&68.75 \\
MaPLe \cite{khattak2023maple}  & 67.8& 68.8& {85.8}& \textbf{92.2}& 50.3   & 76.1& {28.2}& 94.2& 54.7& 71.7 & 68.96\\

TIP-Adapter-F \cite{zhang2022tip}   & 67.1& 67.2&  85.8  & 90.6 &51.6  &{83.8}& 28.8 & 94.0 & \underline{67.8}& 73.4  &71.01 \\

CLIP-Adapter \cite{gao2024clip} & 65.7 & 65.4& 86.1& 89.0& 44.2& 71.3& 25.2& 92.0& 49.3 & 66.9 & 65.51\\

PLOT++ \cite{chen2022prompt}  & \underline{68.8}&
66.8& \underline{86.2}&  {91.9}& \textbf{54.6}&80.5&28.6&\underline{94.3} &65.4&\textbf{74.3} & \underline{71.14} \\

KgCoOp \cite{yao2023visual}  & 66.7&
{68.4}& \textbf{86.4} & \underline{92.1}& 52.7& 74.7&26.8& 94.2& 61.9& 72.8& 69.67\\

ProGrad \cite{zhu2023prompt} & 68.2& 67.0 & 84.9 & 91.4& 52.8 & 80.9 & 28.8& 93.5 & 57.0& 73.3 & 69.78\\

GLoRA \cite{chavan2023one} &7.9 &39.5 &30.9 &68.5 &2.6 &\textbf{87.3} &9.2 &77.2 &12.8 & 41.0 & 37.70\\

TaskRes \cite{yu2023task} &\underline{68.8}& 68.1& 84.6& 90.2& 53.8& 81.7& \textbf{31.3}& 93.6&    65.4& 71.7& 70.92\\

FATE (Ours)      & \textbf{68.9}& \textbf{68.9}& 83.1& 91.2& \underline{53.9} & \underline{84.9}&  \underline{28.9}& \textbf{94.8}& \textbf{73.6} &\underline{73.8} & \textbf{72.20}\\
\bottomrule
\end{tabular}
\end{table*}
\begin{table}
\caption{Comparison of our method with other SSL methods fine-tuned on CLIP across four benchmarks. -V/-T means applying visual/textual prompts as a tuning strategy.}
\label{grip}
\begin{tabular}{lccccc}
\toprule
Method       & DTD   & EuroSAT & Flowers & Aircraft & Avg.   \\
\midrule
CPL-V \cite{zhang2024candidate} & \underline{56.90}  & \textbf{76.96}   & 70.85      & 17.84        & 55.64  \\
FPL-V \cite{zhang2024candidate}  & 51.15 & 66.73   & 69.61      & 19.76        & 51.81  \\
GRIP-V \cite{menghini2023enhancing}   & 54.57 & 63.48   & 67.95      & 19.43        & 51.36  \\
CPL-T \cite{zhang2024candidate}  & \textbf{57.78} & 71.81   & \textbf{84.99}    & 20.59        & \underline{58.79}  \\
FPL-T \cite{zhang2024candidate}  & 51.54 & 63.08   & 74.87      & \underline{20.80}        & 52.57  \\
GRIP-T \cite{menghini2023enhancing}  & 56.07 & 58.66   & 83.60      & 16.98        & 53.83  \\
FATE (Ours)             & 53.93 & \underline{73.60}   & \underline{84.93}      & \textbf{28.87}        & \textbf{60.33}  \\
\bottomrule
\end{tabular}
\end{table}

\begin{table}
\caption{The results for adding DP to the strong augment branch of the implementation on the vision model.}
\label{vit-more-2}
\begin{tabular}{cccccc}
\toprule
  DP  & CIFAR-100 & Semi-iNat & SVHN & SUN397 & CUB200 \\
\midrule
  $\surd$&  48.63& 23.46&  17.10& 37.02& 46.36 \\
 $\times$&  \textbf{59.55}& \textbf{23.86}& \textbf{20.97}& \textbf{40.03}& \textbf{56.96}\\
\bottomrule
\end{tabular}
\end{table}

\subsection{Influence on the Value of $k$}
\label{k-value}
This section will discuss the influence of $k$ on the implementation of the vision-language model. We set $k=1, 2, 4, 8, 16, 32$ and use these pseudo-labeled data sets to train our DP. Furthermore, we conducted ablation experiments on DP and CP, examining the impact of DP and CP on the final predictive performance as $k$ varies. As shown \cref{clip-k}, the prediction effect will be stronger after using DP than the original zero-shot CLIP. Moreover, incorporating DP during the Classification Stage to optimize CP further enhances the model's performance. This shows that when DP is used for the Classification Stage, with the prior knowledge from unlabeled data provided by the DP, the model can perform better in the Classification Stage. When $k$ is very small, the trained DP cannot play a good role, which also indirectly shows that a certain amount of unlabeled data is needed to support the training of DP. When $k$ reaches a certain level, due to the decline in the quality of pseudo-labels, the trained DP will be affected by noise and show a trend of performance convergence or decline.

\subsection{Effectiveness of Fine-tuning with DP}
We conducted more experiments on the vision model to prove the effectiveness of DP as shown in \cref{vit-more-1}. The quality of pseudo-labeling will be improved when the DP is added to pseudo-label unlabeled data in the Classification Stage. We fixed the DP after training it in the Adaptation Stage, and then directly used a labeled sample to fine-tune a classifier and evaluate the model's performance. It was found that the final model performance obtained when adding DP was better than the performance without DP. This shows that our DP learned the overall feature distribution of downstream data and smoothed the semantic information from only one labeled data point to alleviate the overfitting phenomenon. Furthermore, we add DP without updating it to illustrate that DP has indeed learned the feature distribution of downstream data, rather than simply smoothing the classification boundaries of the model by adding noise.

\subsection{Comparison with Other PEFT Strategies}
As for FATE's implementation on the vision-language model, we compare it with a wider range of PEFT strategies in one-shot learning scenarios, including adapter-based approaches: TIP-Adapter-F \cite{zhang2022tip}, CLIP-Adapter \cite{gao2024clip}; alternative prompt tuning methods on CLIP: PLOT++ \cite{chen2022prompt}, KgCoOp \cite{yao2023visual}, ProGrad \cite{zhu2023prompt}, CoOp \cite{zhou2022learning}, CoCoOp \cite{zhou2022conditional}, MaPLe \cite{khattak2023maple}; improved LoRA \cite{hu2022lora} variants adapted for vision tasks: GLoRA \cite{chavan2023one}; and processes that tune CLIP's text embeddings by introducing residual parameters: TaskRes \cite{yu2023task}. Experimental results shown in \cref{PEFT} demonstrate that FATE achieves optimal performance on four benchmarks and suboptimal results on four others among ten evaluated benchmarks, surpassing all other PEFT methods in overall performance. This superior performance indicates that our method exploits additional information from unlabeled data, yielding better results than approaches relying solely on supervised fine-tuning signals.

\subsection{Comparison with Other CLIP-Based SSL Methods}
Furthermore, we compare FATE's implementation on vision-language models against alternative SSL approaches employing prompt tuning on CLIP. These methods \cite{menghini2023enhancing, zhang2024candidate} operate through an iterative process of screening high-confidence samples via CLIP's prompt tuning mechanism and retraining new prompts with the selected samples. While using identical dataset configurations, we maintained all other experimental parameters at their default values as provided in the authors' original implementations. As shown in \cref{grip}, FATE obtains one optimal and two suboptimal results on four benchmarks. Notably, our method outperforms the second-best approach by an average margin of nearly 2 points. This performance gap highlights FATE's superior robustness in extremely label-scarce scenarios, whereas other SSL methods fail to fully leverage this advantage despite utilizing pre-trained models.

\subsection{Discussion on the Position of DP}
We experimented by adding DP to the embedding vector group of the strong augmented view of unlabeled data in the Classification Stage, and found that the performance shown in \cref{vit-more-2} was worse than the final result we provided. It is not necessary to add DP to the branch of the strong augmented view, because the significance of the strong augmented view is to let the model recognize different samples of the same class and reduce the inductive bias of the model. Without DP, the strong augmented view undergoes an additional augmentation at the embedding vector level compared to when DP is included. The reason why the DP is involved in the predicted class distribution mapping of the weak augmented view of the unlabeled data and the labeled data is that we need the predicted class distributions of both to be relatively accurate.

\section{Conclusion}

This paper proposes a two-stage prompt tuning framework called FATE for pre-trained models to address the SSL problem when labeled samples are extremely scarce. FATE is adaptable to both vision and vision-language models. It first uses a large amount of unlabeled data to help the model adapt to the feature distribution of the downstream data, then adapts SSL algorithms to suit pre-trained models, thereby completing the classification task. We demonstrate the effectiveness of FATE through extensive experiments.

\bibliographystyle{ACM-Reference-Format}
\bibliography{sample-base}










\end{document}